\documentclass[10pt,twocolumn,letterpaper]{article}

\usepackage{cvpr}
\usepackage{times}
\usepackage{epsfig}
\usepackage{graphicx}
\usepackage{amsmath}

\usepackage{amssymb}
\newcommand*{\affaddr}[1]{#1} % No op here. Customize it for different styles.
\newcommand*{\affmark}[1][*]{\textsuperscript{#1}}

\usepackage{amsmath}
\usepackage{amssymb}
\usepackage{subfigure}
\usepackage{color}
\usepackage{epsfig}
\usepackage{epstopdf}
\usepackage{algorithm}
\usepackage{algorithmic}
\usepackage{multirow}
\usepackage[table]{xcolor}
\usepackage[pagebackref=true,breaklinks=true,letterpaper=true,colorlinks,bookmarks=false]{hyperref}

 \cvprfinalcopy % *** Uncomment this line for the final submission

\def\cvprPaperID{7204} % *** Enter the CVPR Paper ID here

\ifcvprfinal\fi

\begin{document}

%%%%%%%%% TITLE
\title{SmallBigNet: Integrating Core and Contextual Views for Video Classification}%SmallBig Network for Video Classification
\author{
Xianhang Li\affmark[1]\affmark[2]\footnotemark[1] , 
Yali Wang\affmark[1]\footnotemark[1] , 
Zhipeng Zhou\affmark[1]\footnotemark[1] , and 
Yu Qiao\affmark[1]\affmark[2]\footnotemark[2]\\ 
\\
\affaddr{\affmark[1]ShenZhen Key Lab of Computer Vision and Pattern Recognition, SIAT-SenseTime Joint\\
Lab, Shenzhen Institutes of Advanced Technology, Chinese Academy of Sciences}\\
\affaddr{\affmark[2]SIAT Branch, Shenzhen Institute of Artificial Intelligence and Robotics for Society}
}

\maketitle
 
\renewcommand{\thefootnote}{\fnsymbol{footnote}} 
\footnotetext[1]{Equally-contributed first authors (xianhang710@gmail.com,\\ (yl.wang, zp.zhou)@siat.ac.cn)} 
\footnotetext[2]{Corresponding author (yu.qiao@siat.ac.cn)} 

%\thispagestyle{empty}

%%%%%%%%% ABSTRACT
\begin{abstract}
Temporal convolution has been widely used for video classification.
However,
it is performed on spatio-temporal contexts in a limited view,
which often weakens its capacity of learning video representation.
To alleviate this problem,
we propose a concise and novel SmallBig network,
with the cooperation of small and big views.
For the current time step,
the small view branch is used to learn the core semantics,
while the big view branch is used to capture the contextual semantics.
Unlike traditional temporal convolution,
the big view branch can provide the small view branch with the most activated video features from a broader 3D receptive field.
Via aggregating such big-view contexts,
the small view branch can learn more robust and discriminative spatio-temporal representations for video classification.
Furthermore,
we propose to share convolution in the small and big view branch,
which improves model compactness as well as alleviates overfitting. 
As a result,
our SmallBigNet achieves a comparable model size like 2D CNNs,
while boosting accuracy like 3D CNNs.
We conduct extensive experiments on the large-scale video benchmarks,
e.g.,
Kinetics400, Something-Something V1 and V2.
Our SmallBig network outperforms a number of recent state-of-the-art approaches,
in terms of accuracy and/or efficiency.
The codes and models will be available on \href{https://github.com/xhl-video/SmallBigNet}{https://github.com/xhl-video/SmallBigNet}.

\end{abstract}

\begin{figure*}[t]
\centering
\includegraphics[width=0.99\textwidth]{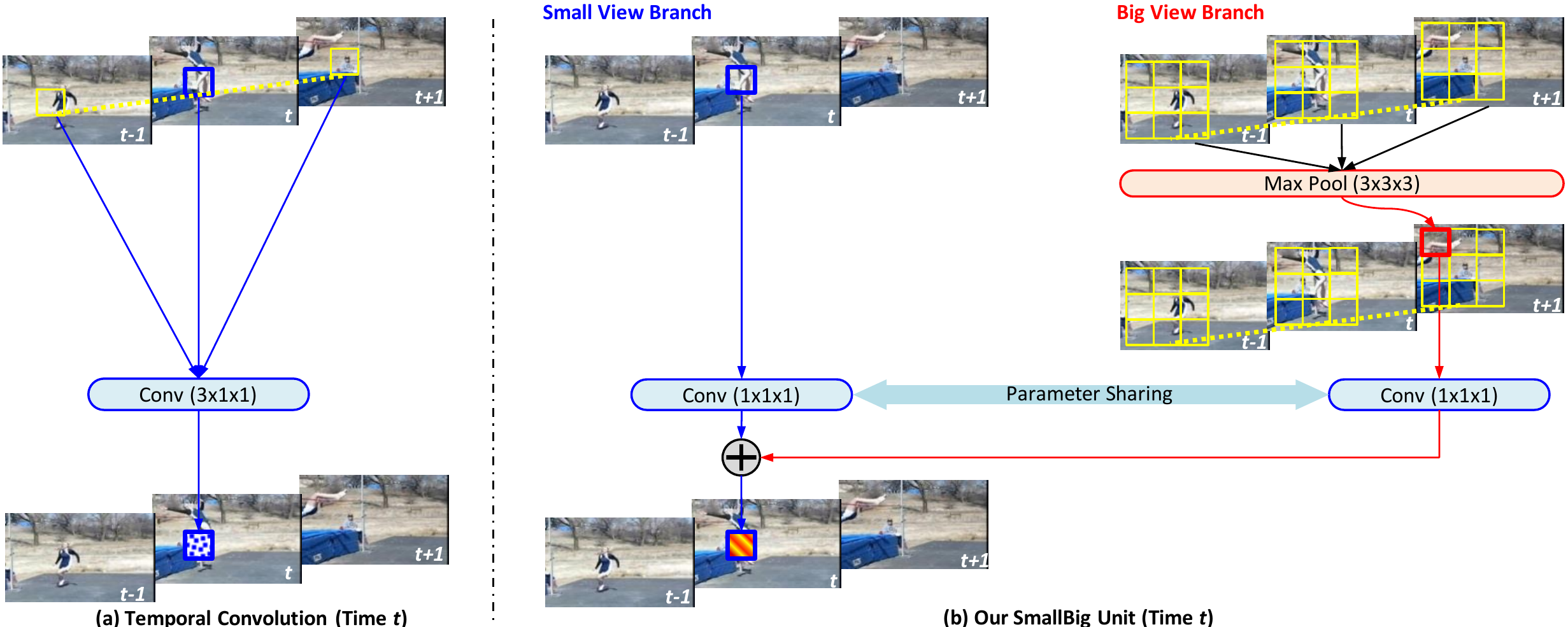}
\caption{Motivation.
As shown in Subplot (a),
temporal convolution is operated on a limited view (yellow tube),
which often contains useless video contexts,
e.g.,
the yellow box at $t+1$ contains the upper body of another sitting person,
without any clues about the moving athlete.
Aggregating such contexts would be harmful to recognize \textit{High Jump}.
To alleviate it,
we propose a novel and concise SmallBig unit with two views,
where
the big view branch can provide the small view branch with the most activated contexts in a broader 3D receptive field.
Such cooperation allows our SmallBig unit to learn more discriminative spatio-temporal representations for video classification.}
\label{fig:Motivations}
\vspace{-0.3cm}
\end{figure*}

\section{Introduction}

3D convolution has been widely used for deep video classification \cite{Carreira_2017_CVPR,Trancvpr2018}.
In particular,
spatio-temporal factorization is preferable to reduce computation cost as well as overfitting \cite{Trancvpr2018,Xie2017}.
However,
temporal convolution in this form is usually operated on a limited view,
which often contains unrelated video contexts.
As shown in Fig.\ref{fig:Motivations}(a),
temporal convolution (e.g., $3\times1\times1$) is performed over the yellow tube of a \textit{High Jump} video.
Apparently,
for the blue box at $t$,
the yellow boxes at $t-1$ and $t+1$ provides almost useless even harmful contexts.
For example,
the yellow box at $t-1$ contains the arm of the athlete.
However,
the arm movement is not quite critical to recognize \textit{High Jump}.
Hence,
the context at $t-1$ tends to be redundant.
Furthermore,
the yellow box at $t+1$ contains the upper body of another sitting person,
without any clues about the moving athlete.
Hence,
the context at $t+1$ tends to be noisy.
By aggregating these contexts with the blue box at $t$,
temporal convolution often leads to a weak and unstable spatio-temporal representation that is not discriminative to recognize \textit{High Jump}.

To tackle the problem above,
we creatively introduce a novel and concise SmallBig unit in Fig.\ref{fig:Motivations}(b),
%we creatively reformulate temporal convolution as the cooperation of two views.
%This eventually produces a novel and concise SmallBig unit in Fig.\ref{fig:Motivations}(b),
where
the big view branch can flexibly provide the small view branch with discriminative contexts from a larger spatio-temporal receptive field.
Via aggregating such contextual clues,
the small view branch tends to learn key spatio-temporal representations for video classification.
Note that,
our SmallBig design is different from the SlowFast design \cite{FeichtenhoferICCV2019},
in terms of both motivation and mechanism.
In particular,
SlowFast is motivated by mimicking two-stream fusion.
Hence,
it feeds input frames of two temporal rates to build up two 3D CNNs (i.e., slow and fast pathways),
and applies lateral connections to integrate them into a unified framework.
Alternatively,
our SmallBig is motivated by releasing the contextual receptive fields of 3D CNN itself.
Hence,
we introduce two distinct views (i.e., small and big branches),
and discover the most activated context of big view to enhance the core representations of small view.

More specifically,
we propose two distinct operations in our SmallBig unit.
First,
we perform 3D max pooling in the big view branch,
which can discover the most activated contexts from a broader 3D tube to enhance the process of the small view branch. 
For example,
we treat the blue box at $t$ as center,
and find its corresponding $3\times3\times3$ yellow tube from $t-1$ to $t+1$ in Fig.\ref{fig:Motivations}(b).
Subsequently,
we apply max pooling over this tube to identify its most activated features,
i.e.,
the red box at $t+1$.
As we can see,
this box contains the take-off pose of the athlete.
Clearly,
it provides more discriminative cues to recognize \textit{High Jump},
comparing to the yellow boxes applied in the limited view of temporal convolution (Fig. \ref{fig:Motivations}(a)).
Second,
we propose to share convolution parameters between the small and big view branches.
This operation improves the compactness of our SmallBig unit,
while boosting accuracy.

Finally,
we build up our SmallBig network (SmallBigNet) in a ResNet style.
By progressively applying a number of SmallBig units in a residual block,
we enlarge the cooperative power of two views with richer contexts from a broader 3D receptive field.
As a result,
our SmallBig network can gradually learn a key spatio-temporal representation for video classification,
when the layer is going deeper.
To evaluate it,
we perform extensive experiments on the widely-used video benchmarks,
e.g.,
Kinetics400 \cite{Kay2017},
Something-Something V1 and V2 \cite{GoyalICCV2017}.
Under the same setting,
our SmallBig network outperforms the recent state-of-the-art methods,
in terms of accuracy and/or efficiency.

\section{Related Works}

\textbf{2D CNNs for Video Classification}.
Over the past years,
video classification has been mainly driven by deep learning frameworks \cite{Carreira_2017_CVPR,FeichtenhoferICCV2019,ref30,Limin2016,Xiaolongcvpr2018,Wu2020DynamicIA}.
One of the widely-used 2D frameworks is two-stream CNNs \cite{ref30},
which can learn video representations respectively from RGB and optical flows.
To further boost performance,
a number of extensions have been proposed by
deep descriptors \cite{ref27},
spatio-temporal fusions \cite{Feichtenhofernips2016,Feichtenhofer2016},
key volume mining \cite{zhucvpr16},
attention \cite{LiminWangcvpr2017},
sequential modeling with RNNs \cite{ref24,ref23,ref15},
temporal segment networks \cite{Limin2016},
temporal relational networks \cite{ZhouECCV2018},
temporal shift module \cite{LinICCV2019},
etc.
In particular,
temporal shift module (TSM) \cite{LinICCV2019} moves feature along the temporal dimension,
which achieves the performance of 3D CNN but maintains the complexity of 2D CNN.
However,
it may lack the comprehensive capacity of understanding spatio-temporal dynamics in videos.
Alternatively,
our SmallBig design can effectively exploit the most-activated contexts from a broader 3D view,
and learn key spatio-temporal representations with cooperation of two different views.
%The experiments show that,
%our SmallBig network can maintain the similar computation efficiency like TSM,
%but achieves a better accuracy.

\textbf{3D CNNs for Video Classification}.
3D CNNs have become popular for spatio-temporal learning,
by treating time as the third dimension of convolutions \cite{Haracvpr2018,ref26,Traniccv2015}.
However,
such operation introduces many more parameters,
which makes 3D convolution harder to train.
To alleviate such problem,
I3D \cite{Carreira_2017_CVPR} has been proposed by inflating 2D convolution into 3D.
But still,
the heavy computation burden limits the power of these full 3D CNNs.
Recent studies have shown that %\cite{Qiuiccv2017,Tranarxiv2019,Trancvpr2018,Xie2017}
factorizing 3D convolution is preferable to reduce complexity as well as boost accuracy,
e.g.,
P3D \cite{Qiuiccv2017},
R(2+1)D \cite{Trancvpr2018},
S3D-G \cite{Xie2017},
etc.
However,
temporal convolution in these approaches is performed on a limited view,
where the unrelated contexts often reduce its capacity of learning video representations.

\textbf{Learning Long-Term Video Dependency}.
Alternatively,
learning long-term dependency has been highlighted for video classification
\cite{ChenCVPR2019,DibaECCV2018,LiuCVPR2019,Xiaolongcvpr2018,WangECCV2018,Wu_2019_ICCV}. %Caoarxiv2019,
%\cite{LiuCVPR2019,Caoarxiv2019,ChenCVPR2019,DibaECCV2018,QiuCVPR2019,Xiaolongcvpr2018}.
One of the most popular models is the nonlocal network \cite{Xiaolongcvpr2018}.
However,
this approach mainly aggregates global relations to assist video classification,
which may not fully exploit fine contexts in the local tube.
%Additionally,
%it introduces extra parameters and thus largely increases computation complexity.
On the contrary,
our approach gradually enlarges contextual receptive fields in a SmallBig block.
Hence,
it allows us to learn key video representation progressively from local view to global view.
%Furthermore,
%our SmallBigNet is much lighter than the nonlocal network,
%by parameter sharing.
%We will further discuss these two networks in the following section.
%The empirical experiments show that,
%it outperforms nonlocal network in terms of both accuracy and efficiency.
Finally,
a SlowFast network has been proposed in \cite{FeichtenhoferICCV2019}.
The key difference is that,
it uses input frames of two temporal rates to mimic two-stream fusion of 3D CNNs,
while
our SmallBig uses two spatio-temporal views on a 3D CNN itself to exploit contexts for enhancing core video features.

\section{SmallBig Unit}

In this section,
we first analyze temporal convolution,
and then explain how to design our SmallBig unit.

\textbf{Temporal Convolution}.
Without loss of generality,
we use a widely-used $3\times1\times1$ temporal convolution filter as illustration.
Specifically,
%we denote $\mathbf{X}_{t}^{(h,w)}\in \mathbb{R}^{C}$ as a $C$-dimension feature vector,
%we denote $\mathbf{x}_{t}^{(h,w)}$ as a feature vector at the spatial location of $(h,w)$ and the temporal location of $t$.
we denote $\mathbf{x}_{t}^{(h,w)}$ as a feature vector at spatial location of $(h,w)$ and temporal frame of $t$.
Additionally,
%we denote $\Theta=[\alpha~~\beta~~\gamma]$ as one filter of this $3\times1\times1$ convolution,
we denote $\Theta=[\Theta_{\alpha},~\Theta_{\beta},~\Theta_{\gamma}]$ as parameters in this $3\times1\times1$ convolution filter.
%where
%%$\alpha$, $\beta$ and $\gamma\in \mathbb{R}^{C}$ are the $C$-dimension parameter vectors that actually refer to the $1\times1\times1$ pointwise convolution.
%$\Theta_{\alpha}$, $\Theta_{\beta}$ and $\Theta_{\gamma}$ actually refer to the $1\times1\times1$ pointwise convolution filter.
As shown in Fig.\ref{fig:Motivations}(a),
temporal convolution applies $\Theta$ to encode video dynamics from $t-1$ to $t+1$,
w.r.t.,
each spatial location $(h, w)$,
\begin{align}
\mathbf{y}_{t}^{(h,w)}&=\text{\textit{\textbf{TemConv}}}(\Theta,~~\mathbf{x}_{t}^{(h,w)},~~\{\mathbf{x}_{t-1}^{(h,w)},~~\mathbf{x}_{t+1}^{(h,w)}\}),\notag\\%\text{TemConv}(\Theta,~~\{\mathbf{X}_{\tau}^{(h,w)}\}_{\tau=t-1}^{t+1}),\notag\\%=\sum\nolimits_{i=(t-1)}^{(t+1)} \mathbf{V}_{i}^{\top}\mathbf{X}_{i}^{(h,w)}
   %&=\underbrace{\mathbf{V}_{t}^{\top}\mathbf{X}_{t}^{(h,w)}}_{small~view }+\underbrace{(\mathbf{V}_{t-1}^{\top}\mathbf{X}_{t-1}^{(h,w)}+\mathbf{V}_{t+1}^{\top}\mathbf{X}_{t+1}^{(h,w)})}_{big~view} \label{eq:TCLX}
&=\Theta_{\beta}\mathbf{x}_{t}^{(h,w)}+[\Theta_{\alpha}\mathbf{x}_{t-1}^{(h,w)}+\Theta_{\gamma}\mathbf{x}_{t+1}^{(h,w)}], \label{eq:TCLX}
%%%%%
%y_{t}^{(h,w)}&=\text{\textit{\textbf{TemConv}}}(\Theta,~~\mathbf{X}_{t}^{(h,w)},~~\{\mathbf{X}_{t-1}^{(h,w)},~~\mathbf{X}_{t+1}^{(h,w)}\}),\notag\\%\text{TemConv}(\Theta,~~\{\mathbf{X}_{\tau}^{(h,w)}\}_{\tau=t-1}^{t+1}),\notag\\%=\sum\nolimits_{i=(t-1)}^{(t+1)} \mathbf{V}_{i}^{\top}\mathbf{X}_{i}^{(h,w)}
%   %&=\underbrace{\mathbf{V}_{t}^{\top}\mathbf{X}_{t}^{(h,w)}}_{small~view }+\underbrace{(\mathbf{V}_{t-1}^{\top}\mathbf{X}_{t-1}^{(h,w)}+\mathbf{V}_{t+1}^{\top}\mathbf{X}_{t+1}^{(h,w)})}_{big~view} \label{eq:TCLX}
%&=\underbrace{\beta\odot\textcolor{blue}{\mathbf{X}_{t}^{(h,w)}}}_{\textcolor{blue}{small~view} }+\underbrace{[\alpha\odot\textcolor{red}{\mathbf{X}_{t-1}^{(h,w)}}+\gamma\odot\textcolor{red}{\mathbf{X}_{t+1}^{(h,w)}}]}_{\textcolor{red}{big ~view}}, \label{eq:TCLX}
\end{align}
where
%$y_{t}^{(h,w)}\in \mathbb{R}$ is the output spatio-temporal representation,
%and
%$\odot$ is the dot-product operation.
$\mathbf{y}_{t}^{(h,w)}$ is the output vector of spatio-temporal representation.
As mentioned in the introduction,
temporal convolution is performed with spatio-temporal contexts of $\mathbf{x}_{t-1}^{(h,w)}$ and $\mathbf{x}_{t+1}^{(h,w)}$.
Such limited view often weakens the discriminative power of $\mathbf{y}_{t}^{(h,w)}$.

%To investigate temporal convolution,
%we propose to reformulate $\mathbf{y}_{t}^{(h,w)}$ as two views in Eq.(\ref{eq:TCLX}).
%The small view branch is used to learn the core semantics at $t$,
%while
%the big view branch is used to capture the contextual semantics from $t-1$ to $t+1$.
%As we can see that,
%the big view branch performs a limited $3\times1\times1$ tube,
%which often weakens the discriminative power of $\mathbf{y}_{t}^{(h,w)}$.

\textbf{SmallBig Unit}.
To address the problem above,
we propose to discover the contexts of $\mathbf{x}_{t}^{(h,w)}$ from a broader spatio-temporal receptive field.
This leads to a novel and concise SmallBig unit with parameters $\Psi=[\Psi_{\rho},~\Psi_{\nu}]$,

\begin{align}
\mathbf{y}_{t}^{(h,w)}
&=\text{\textit{\textbf{SmallBig}}}(\Psi,~~\mathbf{x}_{t}^{(h,w)},~~\{\mathbf{x}_{k}^{(i,j)}\}),\notag\\
&=\underbrace{\Psi_{\rho}\textcolor{blue}{\mathbf{x}_{t}^{(h,w)}}}_{\textcolor{blue}{small~view}}+\underbrace{\Psi_{\nu } \textcolor{red}{\text{MaxPool}(\{\mathbf{x}_{k}^{(i,j)}\})}}_{\textcolor{red}{big~view}}. \label{eq:SmallBig}
%&=\beta\odot[\textcolor{blue}{\mathbf{X}_{t}^{(h,w)}}+\textcolor{red}{\text{MaxPool}(\{\mathbf{X}_{k}^{(i,j)}\})}].
%y_{t}^{(h,w)}
%&=\text{\textit{\textbf{SmallBig}}}(\Psi,~~\mathbf{X}_{t}^{(h,w)},~~\{\mathbf{X}_{k}^{(i,j)}\}),\notag\\
%&=\underbrace{\varphi\odot\textcolor{blue}{\mathbf{X}_{t}^{(h,w)}}}_{\textcolor{blue}{small~view}}+\underbrace{\psi\odot \textcolor{red}{\text{MaxPool}(\{\mathbf{X}_{k}^{(i,j)}\})}}_{\textcolor{red}{big~view}}. \label{eq:SmallBig}
%%&=\beta\odot[\textcolor{blue}{\mathbf{X}_{t}^{(h,w)}}+\textcolor{red}{\text{MaxPool}(\{\mathbf{X}_{k}^{(i,j)}\})}].
\end{align}
%%%%%%%%%%%%%%%%
%\begin{equation}
%y_{t}^{(h,w)}=\underbrace{\beta\odot\textcolor{blue}{\mathbf{X}_{t}^{(h,w)}}}_{\textcolor{blue}{small~view}}+\underbrace{\psi\odot \textcolor{red}{\text{MaxPool}(\{\mathbf{X}_{\tau}^{(i,j)}\}_{\tau=t-1}^{t+1})}}_{\textcolor{red}{big~view}}. \label{eq:SmallBig}
%\end{equation}
Next,
we mainly explain two key operations in this unit.

\textbf{1) 3D Max Pooling Over Big View}.
To further release the spatio-temporal location constraints of contexts,
we propose to work on a broader $T_{\epsilon}\times H_{\epsilon}\times W_{\epsilon}$ tube (e.g., $3\times 3\times 3$) centered at $(t,h,w)$.
In particular,
we perform max pooling over the feature vectors $\{\mathbf{x}_{k}^{(i,j)}\}$ in this 3D tube.
As a result,
we can identify the most activated contextual feature from a bigger view.
Compared to $\mathbf{x}_{t-1}^{(h,w)}$ and $\mathbf{x}_{t+1}^{(h,w)}$ in the temporal convolution,
this max-pooled feature is often more discriminative to capture key video dynamics,
e.g.,
take-off pose of athlete in the red box of Fig.\ref{fig:Motivations}(b).
By aggregating such contexts for $\mathbf{x}_{t}^{(h,w)}$,
our SmallBig unit can reduce redundancy and promote robustness of spatio-temporal learning.

\textbf{2) Parameter Sharing Between Small \& Big Views}.
After obtaining the most activated contextual feature,
%we apply $1\times1\times1$ pointwise convolutions $\varphi$ and $\psi\in \mathbb{R}^{C}$ to further encode the representation in the small and big view branches.
we apply $1\times1\times1$ pointwise convolution filters $\Psi_{\rho}$ and $\Psi_{\nu}$ to further encode the representation in the small and big view branches.
Specifically,
we propose to share the parameters between filters of two views,
i.e.,
$\Psi_{\rho}=\Psi_{\nu}$,
in order to increase model compactness.
Via this operation,
the size of our SmallBig unit is reduced as that of 2D convolution.
%which improves its capacity for deployment in practice.
%Additionally,
%we apply batch normalization individually for each view (right after convolution),
%in order to emphasize different roles of small and big views.
In this case,
our SmallBig unit can efficiently enhance $\mathbf{y}_{t}^{(h,w)}$ with cooperation of two views.

\begin{figure*}[t]
\centering
\includegraphics[width=\textwidth]{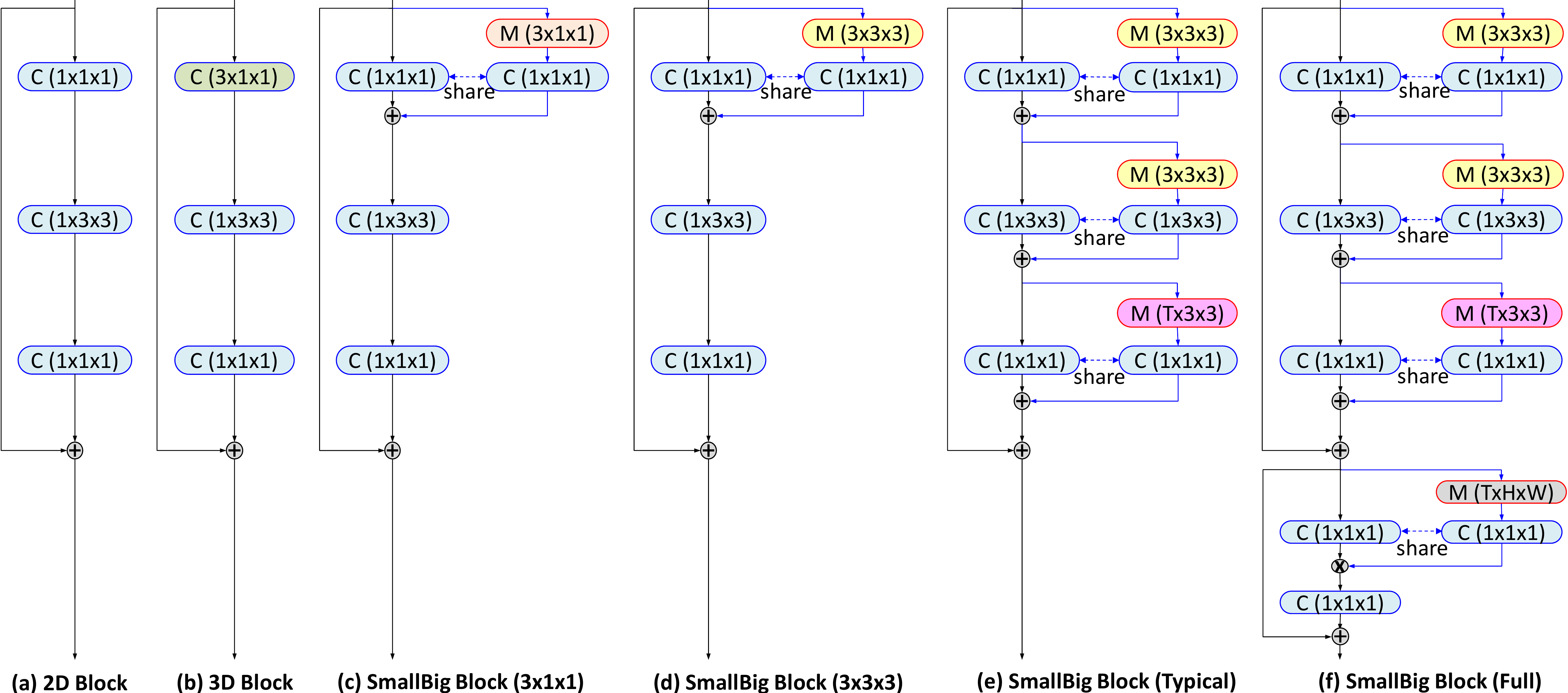}
\caption{SmallBig Blocks.
C: Convolution.
M: Max pooling.
We design a number of SmallBig blocks by adapting all the 2D convolution layers progressively as the SmallBig unit.
More explanations can be found in Section \ref{Exemplar: SmallBig-ResNet}.}
\label{fig:Blocks}
\vspace{-0.3cm}
\end{figure*}

\section{Exemplar: SmallBig-ResNet}
\label{Exemplar: SmallBig-ResNet}

After introducing the SmallBig unit,
we illustrate how to adapt it into a residual style block, %the style of ResNet block,
and then build up a SmallBig network from ResNet23 (or ResNet50).

\textbf{SmallBig Blocks}.
As shown in Fig.\ref{fig:Blocks},
we first introduce two widely-used residual blocks for comparison,
i.e.,
2D convolution in Subplot(a) and 3D convolution in Subplot(b),
where
2D convolution consists of three layers,
i.e., two $1\times1\times1$ and one $1\times3\times3$.
For 3D convolution,
we follow \cite{Xiaolongcvpr2018} to apply inflation \cite{Carreira_2017_CVPR} on the first $1\times1\times1$,
which leads to a $3\times1\times1$ temporal convolution.

For our SmallBig blocks,
we adapt the layers of 2D convolution gradually into the SmallBig unit,
as shown in Subplot(c)-(e).
Note that,
for a typical SmallBig block in Subplot(e),
we set the pooling size as $T\times3\times3$ in the big view branch of last layer,
where $T$ is the total number of sampled video frames.
The main reason is that,
after $3\times3\times3$ max pooling in the big view branch of middle layer,
$1\times3\times3$ convolution further enlarges spatial receptive field.
To balance spatio-temporal view,
we further enlarge temporal receptive field in the big view branch of last layer.

Finally,
we introduce an extra SmallBig unit on top of a typical SmallBig block,
which leads to a full SmallBig block in Subplot(f).
In the big view branch of the extra SmallBig unit,
we operate max pooling over the global spatio-temporal tube of $T\times H\times W$.
In this way,
the full SmallBig block can progressively integrate the most activated contexts from local view to global view.
Furthermore,
the pooling operation in this extra unit actually produces a global feature vector (after conv),
which is irrelevant to spatio-temporal location.
Hence,
%instead of location-wise sum aggregation,
we naturally adapt this vector as attention (with sigmoid),
and apply it for channel-wise product aggregation.
Lastly,
besides of parameter sharing between two views,
we propose to use the bottleneck-like design in this extra SmallBig unit,
e.g.,
input : output channels is 4:1 for its 1st convolution,
while
input : output channels is 1:4 for its 2nd convolution.
This is used to reduce computation cost of our full SmallBig block.

% Please add the following required packages to your document preamble:
% \usepackage{graphicx}
\begin{table}[t]
\centering
\resizebox{0.46\textwidth}{!}{%
\begin{tabular}{c|c|c}
\hline
\multicolumn{2}{c|}{Layer}                   & Output Size \\ \hline
conv1                    & $1\times7\times7$, 64, stride $1\times2\times2$
                         & $8\times112\times112$
                         \\ \hline
pool1                    & $1\times3\times3$ max, stride $1\times2\times2$
                         & $8\times56\times56$
                         \\ \hline
res2                     & $\begin{pmatrix} 1 \times 1 \times 1, 64 \\ 1\times 3 \times 3, 64 \\ 1 \times 1 \times 1, 256 \end{pmatrix}  \times $ 1 (or~3)
                         & $8\times56\times56$
                         \\
                         \hline
res3                     & $\begin{pmatrix} 1 \times 1 \times 1, 256 \\ 1\times 3 \times 3, 256 \\ 1 \times 1 \times 1, 512 \end{pmatrix} \times $ 2 (or~4)
                         & $8\times28\times28$
                         \\
                         \hline
res4                     & $\begin{pmatrix} 1 \times 1 \times 1, 512 \\ 1\times 3 \times 3, 512 \\ 1 \times 1 \times 1, 1024 \end{pmatrix} \times $ 3 (or~6)
                         & $8\times14\times14$
                         \\
                         \hline
res5                     & $\begin{pmatrix} 1 \times 1 \times 1, 1024 \\ 1\times 3 \times 3, 1024 \\ 1 \times 1 \times 1, 2048 \end{pmatrix} \times $ 1 (or~3)
                         & $8\times7\times7$
                         \\
                         \hline
\multicolumn{2}{c|}{global average pool, fc} &  $1\times1\times1$  \\ \hline
\end{tabular}%
}
\caption{2D backbone of our SmallBig network: ResNet23 (or ResNet50).
We construct SmallBig-ResNet by adapting each 2D residual block as a SmallBig block such as Fig.\ref{fig:Blocks}(c)-(f).
The input is $8\times224\times224$,
which is sampled from a 64-frame clip with temporal stride of 8.}
\label{tab:SmallbigResnet}
\vspace{-0.3cm}
\end{table}

\textbf{SmallBig-ResNet}.
After building up the SmallBig blocks,
we construct SmallBig network from ResNet23 (or ResNet50) in Table \ref{tab:SmallbigResnet}.
The size of input clip is $8\times224\times224$.
For each 2D residual block,
we replace it with any of our SmallBig blocks in Fig.\ref{fig:Blocks}(c)-(f).
Note that,
the number of parameters in SmallBig-ResNet is comparable to that of 2D ResNet,
due to parameter sharing.
Moreover,
the parameters of SmallBig-ResNet can be directly initialized from those of 2D ResNet,
which has been well pretrained on ImageNet.
This simplifies the initialization issue and boosts our SmallBig-ResNet in practice.

\section{Further Discussion: SmallBig vs. Nonlocal}
\label{sec:discussion}

As mentioned before,
our SmallBig design is related to the well-known nonlocal operation \cite{Xiaolongcvpr2018},
which also leverages the spatio-temporal contexts in a broader view.
Hence,
we further discuss differences between our design and this SOTA architecture.
For convenience,
we denote $\{\mathbf{x}_{k}^{(i,j)}\}_{all}$ as all the feature vectors in the global tube of $T\times H\times W$.
For $\mathbf{x}_{t}^{(h,w)}$ at $(t,h,w)$,
the nonlocal operation actually finds its highly similar contexts from $\{\mathbf{x}_{k}^{(i,j)}\}_{all}$.
Specifically,
we rewrite this operation as a formulation of two views with parameter $\mathbf{V}=[\mathbf{V}_{\theta},~\mathbf{V}_{\phi},~\mathbf{V}_{g},~\mathbf{V}_{o}]$,
%%%%%%%%%%%%%%%%%%%%%%%%
%\begin{align}
%\mathbf{y}_{t}^{(h,w)}
%&=\text{\textit{\textbf{NonLocal}}}(\mathbf{V},~~\mathbf{x}_{t}^{(h,w)},~~\{\mathbf{x}_{k}^{(i,j)}\}_{all}),\label{eq:nonlocal}\\
%%&=\mathbf{x}_{t}^{(h,w)}+\mathbf{V}_{\varsigma}(\frac{1}{C(\mathbf{x})}\sum\nolimits_{all}f(\mathbf{V}_{\theta}\mathbf{x}_{t}^{(h,w)}, \mathbf{V}_{\phi}\mathbf{x}_{k}^{(i,j)})\mathbf{V}_{g}\mathbf{x}_{k}^{(i,j)})
%%%%%%%%%%%%%%%%%%%
%&=\mathbf{x}_{t}^{(h,w)}+\mathbf{V}_{o}[\sum\nolimits_{all}\frac{f(\mathbf{x}_{t}^{(h,w)}, \mathbf{x}_{k}^{(i,j)})}{C_{x}}g(\mathbf{x}_{k}^{(i,j)})]
%\notag\\
%&=\mathbf{x}_{t}^{(h,w)}+\mathbf{V}_{o}[\sum\nolimits_{all}\frac{f(\mathbf{x}_{t}^{(h,w)}, \mathbf{x}_{k}^{(i,j)})}{C_{x}}\mathbf{V}_{g}\mathbf{x}_{k}^{(i,j)}]
%\notag\\
%&=\underbrace{\textcolor{blue}{\mathbf{x}_{t}^{(h,w)}}}_{\textcolor{blue}{small~view}}+\underbrace{\mathbf{V}_{o}\mathbf{V}_{g}\textcolor{red}{\sum\nolimits_{all}s_{k}^{(i,j)} \mathbf{x}_{k}^{(i,j)}}}_{\textcolor{red}{big~view}}.
%\notag
%\end{align}
%%%%%%%%%%%%%%%%%%%%%%%%
\begin{align}
\mathbf{y}_{t}^{(h,w)}
&=\text{\textit{\textbf{NonLocal}}}(\mathbf{V},~~\mathbf{x}_{t}^{(h,w)},~~\{\mathbf{x}_{k}^{(i,j)}\}_{all})\notag\\
%&=\mathbf{x}_{t}^{(h,w)}+\mathbf{V}_{\varsigma}(\frac{1}{C(\mathbf{x})}\sum\nolimits_{all}f(\mathbf{V}_{\theta}\mathbf{x}_{t}^{(h,w)}, \mathbf{V}_{\phi}\mathbf{x}_{k}^{(i,j)})\mathbf{V}_{g}\mathbf{x}_{k}^{(i,j)})
%%%%%%%%%%%%%%%%%%
&=\mathbf{x}_{t}^{(h,w)}+\mathbf{V}_{o}\sum\nolimits_{all}f(\mathbf{x}_{t}^{(h,w)}, \mathbf{x}_{k}^{(i,j)})g(\mathbf{x}_{k}^{(i,j)})
\notag\\
&=\mathbf{x}_{t}^{(h,w)}+\mathbf{V}_{o}\sum\nolimits_{all}s_{k}^{(i,j)}\mathbf{V}_{g}\mathbf{x}_{k}^{(i,j)}
\notag\\
&=\underbrace{\textcolor{blue}{\mathbf{x}_{t}^{(h,w)}}}_{\textcolor{blue}{small~view}}+\underbrace{\mathbf{V}_{o}\mathbf{V}_{g}\textcolor{red}{\sum\nolimits_{all}s_{k}^{(i,j)} \mathbf{x}_{k}^{(i,j)}}}_{\textcolor{red}{big~view}}.
\label{eq:nonlocal}
\end{align}
$s_{k}^{(i,j)}$ is the similarity score between $\mathbf{x}_{t}^{(h,w)}$ and $\mathbf{x}_{k}^{(i,j)}$.
It is computed from a kernel function,
e.g.,
embedded Gaussian
$f(\mathbf{x}_{t}^{(h,w)}, \mathbf{x}_{k}^{(i,j)})=\exp[(\mathbf{V}_{\theta}\mathbf{x}_{t}^{(h,w)})^{\top}(\mathbf{V}_{\phi}\mathbf{x}_{k}^{(i,j)})]/C_{x}$
with a normalization term $C_{x}$.
Additionally,
$g(\mathbf{x}_{k}^{(i,j)})=\mathbf{V}_{g}\mathbf{x}_{k}^{(i,j)}$ is a linear transformation of $\mathbf{x}_{k}^{(i,j)}$.
Hence,
we move $\mathbf{V}_{g}$ out of summation $\sum$.

Via comparing the big view branch in Eq.(\ref{eq:SmallBig}) and (\ref{eq:nonlocal}),
we find that both mechanisms exhibit the spirit of visual attention.
However,
our SmallBig design contains the distinctive characteristics as follows.
\textbf{First},
the goals of visual attention are different.
The nonlocal operation uses similarity comparison as soft attention,
which aims at finding the similar contexts for $\mathbf{x}_{t}^{(h,w)}$.
Such contexts implicitly assist video classification by modeling spatio-temporal dependency.
Alternatively,
our SmallBig unit uses max pooling as hard attention,
which aims at finding the key contexts around $\mathbf{x}_{t}^{(h,w)}$.
Such contexts are more explicit and discriminative to boost classification accuracy,
since they are highly activated to recognize different video classes.
\textbf{Second},
the receptive fields of visual attention are different.
The nonlocal operation directly works on the global spatio-temporal tube to learn long-term relations,
which may ignore key video details for classification.
Alternatively,
our SmallBig unit works on the local spatio-temporal tube to capture fine video clues.
More importantly,
we gradually enlarge the receptive field of big view branch in the SmallBig block,
allowing us to learn video representation progressively from local view to global view.
%\textbf{Finally},
%the nonlocal operation introduces extra parameters,
%while our SmallBig unit inherits the parameters of 2D CNN and maintains the model size by parameter sharing.
%Hence,
%our SmallBig network is more light-weight in practice.
Our experiments also show that
the SmallBig network steadily outperforms the nonlocal network.

\begin{table*}[t]
\centering
%%%%%%%%%%%%%
\subtable{
\begin{minipage}[t]{0.28\textwidth}
\centering
\resizebox{\textwidth}{!}{%
\begin{tabular}{l|cc}
\hline
R23                             & Top1 & Top5  \\ \hline
2D                              &   64.1   &  85.4     \\
3D: 3$\times$1$\times$1         &   68.3   &  88.2     \\
SmallBig: 3$\times$1$\times$1   &   \textbf{69.0}   &  \textbf{88.6}     \\
\hline
\end{tabular}%
%\begin{tabular}{l|cc|cc}
%\hline
%R23                             & Params & GFlops  & Top1 & Top5  \\ \hline
%2D                              &  11.3M     &  17       &   64.1   &  85.4     \\
%3D: 3$\times$1$\times$1         &            &           &   68.3   &  88.2     \\
%SmallBig: 3$\times$1$\times$1   &  11.3M     &  21       &   \textbf{69.0}   &  \textbf{88.6}     \\
%\hline
%\end{tabular}%
}
\caption{Effectiveness of SmallBig.
We apply ResNet23 (R23) as 2D backbone,
and adapt 3D block ($3\times1\times1$) of Fig.\ref{fig:Blocks}(b) and SmallBig block ($3\times1\times1$) of Fig.\ref{fig:Blocks}(c) in all the residual stages.}
\label{tab:effect}
\end{minipage}
}
\hspace{0.3cm}
%%%%%%%%%%%
\subtable{
\begin{minipage}[t]{0.31\textwidth}
\centering
\resizebox{\textwidth}{!}{%
\begin{tabular}{l|cc}
\hline
SmallBig-R23                       & Top1 & Top5 \\ \hline
R23                                & 64.1 & 85.4 \\ \hline
SmallBig: res2+3+4+5               & \textbf{69.0} & \textbf{88.6} \\
SmallBig: res3+4+5                 &  68.8    &    88.5       \\
SmallBig: res4+5                   &  68.6    &    88.5         \\
SmallBig: res5                     &  64.5    &    85.6           \\
%0-block: 2D-R23           &      &      \\ \hline
%1-block: res2         &      &      \\
%1-block: res5        &      &      \\
%2-block: res2+5         &      &      \\
%2-block: res3         &      &      \\
%3-block: res2+3         &      &      \\
%3-block: res3+5        &      &      \\
%3-block: res4        &      &      \\
%4-block: res2+3+5        &      &      \\
%4-block: res2+4        &      &      \\
%4-block: res4+5        &      &      \\
%5-block: res2+4+5         &      &      \\
%5-block: res3+4         &      &      \\
%6-block: res2+3+4         &      &      \\
%6-block: res3+4+5         &      &      \\
%7-block: res2+3+4+5         &      &      \\
\hline
\end{tabular}%
}
\caption{Stage of using SmallBig blocks.
As expected,
the middle blocks (e.g., res4) are often more important than others.}
\label{tab:stage}
\end{minipage}
}
\hspace{0.3cm}
%%%%%%%%%%%
\subtable{
\begin{minipage}[t]{0.28\textwidth}
\centering
\resizebox{\textwidth}{!}{%
\begin{tabular}{l|cc}
\hline
SmallBig-R23                        & Top1 & Top5 \\ \hline
R23                              & 64.1 & 85.4    \\ \hline
SmallBig: 3$\times$1$\times$1    & 69.0 & 88.6    \\
SmallBig: 3$\times$3$\times$3    & \textbf{69.5} & \textbf{89.0}    \\
SmallBig: 3$\times$5$\times$5    & 69.1 & 88.5    \\
SmallBig: 3$\times$7$\times$7    & 68.6 & 88.3    \\
SmallBig: T$\times$3$\times$3    & 69.3 & 88.5    \\
\hline
%First: $3\times1\times1$         &      &                         \\
%First: $3\times3\times3$         &      &               \\
%First: $3\times5\times5$         &      &               \\
%First: $3\times7\times7$         &      &               \\ \hline
%Last: $T\times1\times1$         &      &                         \\
%Last: $T\times3\times3$         &      &               \\
%Last: $T\times5\times5$         &      &               \\
%Last: $T\times7\times7$         &      &               \\ \hline
\end{tabular}%
}
\caption{Broader receptive field in the big view branch.}
\label{tab:recep}
\end{minipage}
}
%%%%%%%%%%%
\subtable{
\begin{minipage}[t]{0.5\textwidth}
\centering
\resizebox{\textwidth}{!}{%
\begin{tabular}{l|cc}
\hline
SmallBig-R23                        & Top1 & Top5 \\ \hline
R23                              & 64.1 & 85.4                       \\ \hline
SmallBig: 3$\times$3$\times$3(1st layer)    & 69.5 & 89.0                        \\
SmallBig: 3$\times$3$\times$3(1st layer)+T$\times$3$\times$3(3rd layer) & 70.8     & 89.3              \\
SmallBig: Typical        &  71.4    &  90.0             \\
SmallBig: Full         & \textbf{72.6}     &  \textbf{90.3}             \\ \hline
\end{tabular}%
}
\caption{More SmallBig layers.
The accuracy is consistently better,
when more layers are progressively changed as our SmallBig design.
For SmallBig: Typical or Full,
all blocks refer to Fig.\ref{fig:Blocks}(e) or (f).}
\label{tab:layers}
\end{minipage}
}
\hspace{0.3cm}
%%%%%%%%%%%
\subtable{
\begin{minipage}[t]{0.40\textwidth}
\centering
\resizebox{\textwidth}{!}{%
\begin{tabular}{l|cc|cc}
\hline
SmallBig-R23           & Params    & GFlops & Top1   & Top5     \\ \hline
R23                    & 11.3M     &  17    & 64.1              &  85.4     \\ \hline
Avg Pool               & 13.4M     &  31    & 72.2              &  90.0        \\
Max Pool               & 13.4M     &  31    & \textbf{72.6}     &  \textbf{90.3}      \\ \hline
Without Share          & 22.1M     &  31    & 71.6              &  89.6        \\
With Share             & 13.4M     &  31    & \textbf{72.6}     &  \textbf{90.3}      \\ \hline
Single BN              & 13.3M     &  17    & 64.6              &  85.8        \\
Individual BN          & 13.4M     &  31    & \textbf{72.6}     &  \textbf{90.3}      \\
\hline
\end{tabular}%
}
\caption{Detailed designs of SmallBig.}
\label{tab:detail}
\end{minipage}
}
\hspace{0.3cm}
%%%%%%%%%%%
\subtable{
\begin{minipage}[t]{0.4\textwidth}
\centering
\resizebox{\textwidth}{!}{%
\begin{tabular}{l|cc|cc}
\hline
Model                             & Params & GFlops  & Top1 & Top5  \\ \hline
R23                                 & 11.3M     &  17    & 64.1              &  85.4       \\
NonLocal-R23                        & 18.7M     &  34    & 70.2              &  89.1     \\
SmallBig-R23                        & 13.4M     &  31    & \textbf{72.6}     &  \textbf{90.3}     \\
\hline
\end{tabular}%
}
\caption{SmallBig vs. NonLocal.
Our SmallBig-R23 outperforms NonLocal-R23,
showing the superiority of our SmallBig design when finding contexts.}
\label{tab:non}
\end{minipage}
}
\hspace{0.3cm}
%%%%%%%%%%%
%\subtable{
%\begin{minipage}[t]{0.32\textwidth}
%\centering
%\resizebox{\textwidth}{!}{%
%\renewcommand\tabcolsep{3pt}
%\begin{tabular}{l|cc|l|cc}
%\hline
%Backbone                            & Top1 & Top5 &Backbone                          & Top1 & Top5 \\ \hline
%R23                              &      &          & R50                             &      &             \\
%SmallBig-R23                     &      &          & SmallBig-R50                    &      &                         \\ \hline
%\end{tabular}%
%}
%\makeatletter\def\@captype{table}\makeatother\caption{2}
%\end{minipage}
%}
%%%%%%%%%%%%
\subtable{
\begin{minipage}[t]{0.25\textwidth}
\centering
\resizebox{\textwidth}{!}{%
\begin{tabular}{l|cc}
\hline
Backbone                            & Top1 & Top5   \\ \hline
R23                              & 64.1              &  85.4                  \\
SmallBig-R23                     & 72.6              &  90.3                  \\ \hline
R50                              & 70.4              &  89.1                  \\
SmallBig-R50                     & \textbf{76.3}     &  \textbf{92.5}         \\ \hline
\end{tabular}%
}
\caption{Backbone.
Our SmallBig-R23 even outperforms R50.}
\label{tab:deep}
\end{minipage}
}
\hspace{0.3cm}
%%%%%%%%%%%
\subtable{
\begin{minipage}[t]{0.24\textwidth}
\centering
\resizebox{\textwidth}{!}{%
\renewcommand\tabcolsep{2pt}
\begin{tabular}{l|cc}
\hline
SmallBig-R50                 & Top1              & Top5           \\ \hline
Extra Unit: Simple         & 75.8              & 92.1           \\
Extra Unit: Default         & \textbf{76.3}     & \textbf{92.5}  \\ \hline
\end{tabular}%
}
\caption{Extra unit in SmallBig (Full).
For comparison,
we replace the extra unit in Fig.\ref{fig:Blocks}(f) by a simplified version.}
\label{tab:attention}
%\resizebox{\textwidth}{!}{%
%%\renewcommand\tabcolsep{3pt}
%\begin{tabular}{l|cc}
%\hline
%SmallBig-R50                 & Top1 & Top5   \\ \hline
%8-frame                      & 76.3 &  92.5    \\
%16-frame                     &      &        \\ \hline
%\end{tabular}%
%}
%\caption{More video frames.
%The 8/16-frame input is sampled from a 64-frame clip with temporal stride of 8/4.}
%\label{tab:frame}
\end{minipage}
}
%%%%%%%%%%%
\vspace{-0.5cm}
\end{table*}

\section{Experiment}

\textbf{Data Sets}.
We perform the experiments on the large-scale video benchmarks,
i.e.,
Kinetics400 \cite{Kay2017},
Something-Something V1 and V2 \cite{GoyalICCV2017}.
Kinetics400 consists of around 300k videos from 400 categories.
Something-Something V1/V2 consists of around 108k/220k videos from 174 categories.
We mainly evaluate all the models on the validation set,
where
we report Top1 \& Top5 accuracy (\%) and GFlops to comprehensively evaluate accuracy and efficiency.

\textbf{Training}.
For all the data sets,
we follow \cite{Xiaolongcvpr2018} to use the spatial size of $224\times224$,
which is randomly cropped from a scaled video whose shorter side is randomly sampled in $[256,320]$ pixels.
For Kinetics400,
the input clip consists of 8 frames,
which are sampled from 64 consecutive frames with temporal stride 8.
We train our models with 110 epochs,
where
we set the weight decay as 1e-4,
and
utilize the cosine schedule of learning rate decay.
For SmallBig-ResNet23,
we set the initial learning rate as 0.02 and the batch size as 128.
For SmallBig-ResNet50/101,
we set the initial leaning rate as 0.00625 and the batch size as 128.
For Something-Something V1 and V2,
we divide a video into 8 segments and then randomly choose one frame in each segment.
We train our models with 50 epochs.
The initial learning rate is 0.01,
and it decays at 30, 40, 45 epochs respectively.
Finally,
we apply batch normalization individually for each view (right after convolution).
All the models are pretrained on ImageNet,
including BN in the small view branch of each layer.
For BN in the big view branch,
we initialize its scale parameter as zero.
This design makes the initial state of our SmallBig network as the original ResNet.

\textbf{Inference}.
Following \cite{FeichtenhoferICCV2019,Xiaolongcvpr2018},
we rescale the video frames with the shorter side 256 and take three crops (left, middle, right) of size $256\times256$ to cover the spatial dimensions.
%Then,
%we resize these frames as $224\times224$.
Unless stated otherwise,
we uniformly sample 10/2 clips for Kinetics400 / Something-Something V1 and V2.
We average their softmax scores for video-level prediction.

\subsection{Evaluation on Kinetics400}

In the following,
we perform extensive ablation studies to investigate various distinct characteristics in our SmallBig network.
Then,
we further evaluate accuracy and efficiency of our SmallBig networks by a comprehensive comparison with the recent state-of-the-art approaches.

\textbf{Effectiveness of SmallBig}.
We apply ResNet23 (R23) of Table \ref{tab:SmallbigResnet} as backbone,
and adapt the SmallBig block ($3\times1\times1$) of Fig.\ref{fig:Blocks}(c) into all the residual stages.
For comparison,
we also adapt the 3D block ($3\times1\times1$) of Fig.\ref{fig:Blocks}(b) in the same way.
As shown in Table \ref{tab:effect},
our SmallBig-R23 outperforms its 2D and 3D counterparts.
Note that,
even though we do not further enlarge the spatio-temporal receptive field in the big view branch,
our SmallBig block ($3\times1\times1$) still achieves a better result than the 3D block ($3\times1\times1$).
It illustrates that,
the most activated contexts found by max pooling is a preferable guidance to learn key video representation,
compared to temporal convolution.

\textbf{Stage of using SmallBig blocks}.
We use the above SmallBig-R23 ($3\times1\times1$) to evaluate which stage may be important for SmallBig design.
In Table \ref{tab:stage},
we gradually recover our SmallBig blocks as the original 2D blocks,
from bottom to top.
As expected,
the middle blocks (e.g., res4) are often more important than the bottom and top blocks.
The main reason is that,
the receptive field is too small (or big) in the bottom (or top) blocks,
enlarging 3D view tends to find useless (or similar) contexts.
On the contrary,
the middle blocks contain the middle-level semantics with a reasonable spatio-temporal receptive field.
Hence,
the contexts in these blocks would be more discriminative.
In our following experiments,
we use SmallBig blocks in all the residual stages to achieve the best accuracy.

\textbf{Broader receptive field in the big view branch}.
For SmallBig-R23 ($3\times1\times1$),
we further extend 3D receptive field in its big view branch.
As shown in Table \ref{tab:recep},
the accuracy first increases and then decreases.
It may be because the diversity of contexts is reduced,
when we directly perform max pooling on too big view.
As a result,
SmallBig-R23 achieves the best performance with $3\times3\times3$ in Fig.\ref{fig:Blocks}(d).

\textbf{More SmallBig layers}.
The above experiment in Table \ref{tab:recep} indicates that,
it is not reasonable to enlarge the 3D receptive field directly to a very big view.
Hence,
we adapt more layers in each residual block to be our SmallBig unit,
allowing to extend the 3D receptive field gradually from local to global view.
As shown in Table \ref{tab:layers},
the accuracy is consistently getting better,
when more convolution layers are progressively changed as SmallBig.
As expected,
the setting of SmallBig-R23 (Full) achieves the best performance,
where all the SmallBig blocks refer to Fig.\ref{fig:Blocks}(f).
In the following,
we use the full setting in our experiments.

\textbf{Detailed designs of SmallBig}.
We use SmallBig-R23 (Full) to further investigate the detailed designs of SmallBig in Table \ref{tab:detail}.
\textbf{Avg Pool vs Max Pool}.
We apply different pooling operations in the big view branch.
The performance of max pooling is better.
Hence,
we choose it in our experiments.
\textbf{Without Sharing vs With Sharing}.
We apply different parameter sharing strategies for convolution in our SmallBig unit.
As expected,
parameter sharing can reduce the model size of our SmallBig-R23 as the original R23.
Its accuracy is even slightly better than the non-sharing case.
Hence,
we share parameters between small and big view branches.
\textbf{Single BN vs Individual BN}.
As mentioned in the implementation details,
we apply BN individually for each view,
i.e.,
BN(conv(small))+BN(conv(big)).
This would introduce extra flops.
To further reduce complexity,
we apply a single BN directly on the output representation,
i.e.,
BN(conv(small)+conv(big)).
Due to the linearity of convolution and sum operations,
this operation is equivalent to BN(conv(small+big)),
which only requires a single convolution and decreases flops as 2D CNN.
As shown in Table \ref{tab:detail},
single BN has higher efficiency but much lower accuracy,
while
individual BN has higher accuracy but lower efficiency.
For consistency,
we choose Individual BN to achieve a better accuracy.

\textbf{SmallBig vs. NonLocal}.
We compare our SmallBig design with the related NonLocal operation.
Specifically,
we use a preferable setting of NonLocal as suggestion in \cite{Xiaolongcvpr2018},
where
NL is added on all the residual blocks of res3 and res4.
As shown in Table \ref{tab:non},
our SmallBig-R23 outperforms NonLocal-R23.
It illustrates that,
to boost performance,
it is preferable to find the most activated contexts progressively from local to global view,
instead of finding dependent contexts directly on the global view.

\textbf{Deeper backbone}.
We further investigate the performance of our SmallBig network,
with a deeper backbone,
e.g.,
ResNet50 (R50).
As shown in Table \ref{tab:deep},
our SmallBig-R23 even outperforms R50.
It illustrates the power of our SmallBig design.
Furthermore,
SmallBig-R50 outperforms SmallBig-R23,
showing the effectiveness of SmallBig in deeper backbones.

\textbf{Extra unit in SmallBig (Full)}.
As shown in Fig.\ref{fig:Blocks}(f),
we add an extra SmallBig unit with global pooling.
Note that,
we apply channel-wise product aggregation in the extra unit.
Hence,
we design a Squeeze-and-Excitation version for comparison.
%as a simplified version(SE) \cite{SENet}
Specifically,
we first perform global spatio-temporal pooling,
and then add two extra $1\times1\times1$ convolutions.
The resulting vector (after sigmoid) is used as channel-wise attention for residual aggregation.
In Table \ref{tab:attention},
our default design outperforms this simplified design in the extra unit.
It illustrates that,
our default extra unit is a preferable choice of global spatio-temporal aggregation,
with cooperation of two views.

\textbf{Comparison with the SOTA approaches}.
We make a comprehensive comparison in Table \ref{tab:kSOTA},
where
our SmallBig network outperforms the recent SOTA approaches.
\textbf{First},
our 8-frame SmallBig-R50 outperforms 32-frame Nonlocal-R50 \cite{Xiaolongcvpr2018} (Top1 acc: 76.3 vs. 74.9),
and it uses $4.9\times$ less GFlops than 128-frame Nonlocal-R50 but achieves a competitive accuracy  (Top1 acc: 76.3 vs. 76.5).
Moreover,
it is even slightly better than 32-frame Nonlocal-R101 (Top1 acc: 76.3 vs. 76.0).
All these results clearly illustrate that,
our SmallBig network is a more accurate and efficient approach than the nonlocal network,
for modeling contexts in video classification.
\textbf{Second},
with the comparable GFlops,
our 8-frame SmallBig-R50 outperforms 36-frame SlowFast-R50 \cite{FeichtenhoferICCV2019} (Top1 acc: 76.3 vs. 75.6).
It indicates the importance of SmallBig in context exploitation of 3D CNN itself.
Additionally,
we perform score fusion over 8-frame SmallBig-R50 and 32-frame SmallBig-R101,
which mimics two-steam fusion with two temporal rates.
When testing,
we use 4 clips and 3 crops per clip to maintain computation.
Our SmallBig$_{En}$ achieves a better accuracy than SlowFast,
using the same number of frames.
\textbf{Finally},
our 8-frame SmallBig-R50 outperforms 8-frame TSM-R50 \cite{LinICCV2019} (Top1 acc: 76.3 vs. 74.1).
It shows that,
spatio-temporal learning of SmallBig is more effective than temporal shift of TSM.

\begin{table}[t]
\centering
\resizebox{0.48\textwidth}{!}{
\begin{tabular}{l|l|l|cc|l}
  \hline\hline
  % after \\: \hline or \cline{col1-col2} \cline{col3-col4} ...
  Method                                  & Backbone        & Frame, Size                & Top1    & Top5      & GFlops$\times$crops   \\
  \hline\hline
  STC\cite{DibaECCV2018}                  & ResNeXt101  & 32, 112          & 68.7    & 88.5      & N/A$\times$N/A          \\
  ARTNet\cite{ARTNet}                     & R18         & 16, 112          & 69.2    & 88.3      & 5,875=23.5$\times$250        \\
  MFNet\cite{ChenECCV2018}                & R34     & 16, 224          & 72.8    & 90.4      & 11$\times$N/A           \\
  R(2+1)D\cite{Trancvpr2018}              & R34     & 8, 112           & 74.3    & 91.4      & 152$\times$N/A         \\
  I3D\cite{Carreira_2017_CVPR}           & Inception  & 64, 224          & 71.1    & 89.3      & 108$\times$N/A       \\
  S3D-G\cite{Xie2017}                  & Inception  & 64, 224          & 74.7    & 93.4      & 71.4$\times$N/A         \\
  \hline\hline
  A$^{2}$-Net\cite{ChenNeuIPS2018}        & R50     & 8, 224           & 74.6    & 91.5      & 41$\times$N/A          \\
  SlowOnly\cite{FeichtenhoferICCV2019}    & R50     & 8, 224           & 74.9    & 91.5      & 1,257=41.9$\times$30          \\
  GloRe\cite{ChenCVPR2019}                & R50     & 8, 224           & 75.1    & N/A       & 867=28.9$\times$30          \\
% GCNet\cite{Caoarxiv2019}                & R50     & 8, 224           & 76.0    & 92.3      & 39.4$\times$30           \\
  TSM\cite{LinICCV2019}                   & R50     & 8, 224           & 74.1    & 91.2      & 990=33$\times$30          \\
TEI\cite{Liu2019TEINetTA}                   & R50     & 8, 224          & 74.9    & 91.8       & 990=33$\times$30          \\
  
  TSM\cite{LinICCV2019}                   & R50     & 16, 224          & 74.7    & N/A       & 1,950=65$\times$30          \\
  TEI\cite{Liu2019TEINetTA}                   & R50     & 16, 224          & 76.2    & 92.5       & 1,980=66$\times$30          \\
  %\rowcolor{gray!10}
  SlowFast\cite{FeichtenhoferICCV2019}    & R50+R50  & 36=4+32, 224     & 75.6    & 92.1   & 1,083=36.1$\times$30         \\%\rowcolor{gray!10}
%  SlowFast\cite{FeichtenhoferICCV2019}    & R50         & 40=8+32, 224     & 77.0    & 92.6      & 65.7$\times$30           \\
  NL I3D\cite{Xiaolongcvpr2018}           & R50     & 32, 224          & 74.9    & 91.6      & N/A$\times$N/A          \\%\rowcolor{gray!10}
  NL I3D\cite{Xiaolongcvpr2018}           & R50     & 128, 224         & 76.5    & 92.6      & 8,460=282$\times$30           \\%\rowcolor{gray!10}
  \hline
  Our SmallBig                            & R50     & 8, 224           & \textbf{76.3}    & \textbf{92.5}      & 1,710=57$\times$30           \\
  \hline\hline
  CoST\cite{LiCVPR2019}                   & R101    & 8, 224           & 75.5    & 92.0      & N/A$\times$N/A           \\
  GloRe\cite{ChenCVPR2019}                & R101    & 8, 224           & 76.1    & N/A       & 1,635=54.5$\times$30          \\
  CPNet\cite{LiuCVPR2019}                 & R101    & 32, 224          & 75.3    & 92.4      & N/A$\times$N/A         \\%\rowcolor{gray!10}
  NL I3D\cite{Xiaolongcvpr2018}           & R101    & 32, 224          & 76.0    & 92.1      & N/A$\times$N/A          \\ %\rowcolor{gray!10}
  NL I3D\cite{Xiaolongcvpr2018}           & R101    & 128, 224         & 77.7    & 93.3      & 10,770=359$\times$30            \\%\rowcolor{gray!10}
%  CoST\cite{LiCVPR2019}                   & R101    & 32, 224          & 77.5    & 93.2      & N/A$\times$N/A           \\
%  ip-CSN\cite{Tranarxiv2019}              & R101    & 32, 224          & 78.5    & 93.5      & 83$\times$30            \\
%  CorrNet\cite{Tranarxiv2019corre}        & R101        & 32, 224          & 78.5    & N/A       & 224$\times$30            \\
  SlowFast\cite{FeichtenhoferICCV2019}    & R101+R101   & 40=8+32, 224     & 77.9    & 93.2      & 3,180=106$\times$30           \\ %\rowcolor{gray!10}
  SlowFast\cite{FeichtenhoferICCV2019} & R101+R101        & 80=16+64, 224    & 78.9    & 93.5      & 6,390=213$\times$30            \\
%  LGD-3D\cite{QiuCVPR2019}                & R101    & 128, 112         & 79.4    & 94.4      & N/A $\times$N/A          \\
%  \hline
%  ip-CSN\cite{Tranarxiv2019}              & R152   & 32, 224          & 79.2    & 93.8      & 108.8$\times$30        \\
  \hline
  Our SmallBig                             & R101     & 32, 224           &   77.4  & 93.3        & 5,016=418$\times$12     \\ %\rowcolor{gray!10}
  Our SmallBig$_{En}$                      & R50+R101 & 40=8+32, 224      &   \textbf{78.7}  & \textbf{93.7}        & 5,700=475$\times$12     \\
  \hline\hline
\end{tabular}
}
\caption{Comparisons with SOTA on Kinetics400 validation set (RGB input).
Our 8-frame SmallBig-R50 outperforms 32-frame Nonlocal-R50 with a higher accuracy,
and uses $4.9\times$ less GFlops than 128-frame Nonlocal-R50 but with a competitive accuracy.
Its accuracy is even slightly better than 32-frame Nonlocal-R101.
Moreover,
with the comparable GFlops,
our 8-frame SmallBig-R50 outperforms 36-frame SlowFast-R50.
All these results show that,
our SmallBig network is an accurate and efficient model for video classification.}
\label{tab:kSOTA}
\vspace{-0.3cm}
\end{table}

\begin{table}[t]
\centering
\resizebox{0.48\textwidth}{!}{
\renewcommand\tabcolsep{3pt}
\begin{tabular}{l|l|l|cc|cc|cc}
  \hline\hline
  % after \\: \hline or \cline{col1-col2} \cline{col3-col4} ...
  \multirow{2}{*}{\small{Method}} & \multirow{2}{*}{\small{Backbone}} & \multirow{2}{*}{\small{Frame,1Clip,1Crop}} & \multicolumn{2}{c|}{\small{V1}} & \multicolumn{2}{c|}{\small{V2}} & \multirow{2}{*}{\small{GFlops}}  \\
                          &                           &                  &   \small{Top1}    &    \small{Top5}   &     \small{Top1}    &   \small{Top5}   &                          \\
  \hline\hline
  TSN            \cite{ZhouECCV2018}                   & Inception        & 8     &  19.5  & -    &  -   &  -  & 16      \\
  TRN$_{multiscale}$ \cite{ZhouECCV2018}               & Inception        & 8    &  34.4  & -    &  -   &  - & 16          \\
%  TRN-2Stream  \cite{ZhouECCV2018}                    & Inception        & 8+8   &  42.0  & -    &  -   &  - & -        \\
  ECO  \cite{ZolfaghariECCV2018}                       & Incep+R18     & 8      & 39.6  & -    &  -   & - &  32     \\
  ECO  \cite{ZolfaghariECCV2018}                       & Incep+R18     & 16     & 41.4  & -    &  -   & - &  64      \\
  ECO$_{En}$Lite  \cite{ZolfaghariECCV2018}            & Incep+R18     & 92      & 46.4  & -    &  -   & - &  267    \\
%  ECO$_{En}$Lite-2Stream  \cite{ZolfaghariECCV2018}  & Incep+R18     & 184=92+92   & 49.5  & -    &  -   & - &  -     \\
  TSM  \cite{LinICCV2019}                              & R50               & 8                & 45.6    & 74.2& -    &  -  &  33   \\
  TSM  \cite{LinICCV2019}                              & R50               & 16               & 47.2    & 77.1& -    &  -  &  65    \\
  TSM$_{En}$  \cite{LinICCV2019}                       & R50               & 24=8+16          & 49.7    & 78.5& -    &  - &  98    \\
%  TSM-2Stream  \cite{LinICCV2019}                      & R50              & 32=16+16         & 52.6    & 81.9 & -    &  -  &  -     \\
  \hline
  Our SmallBig                        & R50            & 8                & 47.0   & 77.1  & 59.7   &  86.7   &  52      \\
  Our SmallBig                        & R50            & 16               & 49.3   & 79.5  & 62.3     & 88.5    &  105       \\
  Our SmallBig$_{En}$                 & R50            & 24=8+16          & \textbf{50.4}   & \textbf{80.5}  & \textbf{63.3}     & \textbf{88.8}    &  157      \\
 \hline\hline
  % after \\: \hline or \cline{col1-col2} \cline{col3-col4} ...
    \multirow{2}{*}{\small{Method}} & \multirow{2}{*}{\small{Backbone}} & \multirow{2}{*}{\small{Frame$\times$Clip$\times$Crop}} & \multicolumn{2}{c|}{\small{V1}} & \multicolumn{2}{c|}{\small{V2}} & \multirow{2}{*}{\small{GFlops}}  \\
                          &                           &                  &   \small{Top1}    &    \small{Top5}   &     \small{Top1}    &   \small{Top5}   &                          \\
 \hline\hline
%  TSN               \cite{LinICCV2019}                 & Inception        & 8     &  -  & -    &  30.0  &  60.5  & -       \\
%  TRN$_{multiscale}$ \cite{ZhouECCV2018}               & Inception        & 8     &  -  & -    &  48.8  &  77.6 & -         \\
  I3D            \cite{WangECCV2018}                   & R50           & 64=32$\times$2     &  41.6    &  72.2& -    &  - &        \\
  NL I3D         \cite{WangECCV2018}                   & R50           & 64=32$\times$2     &  44.4    &  76.0& -    &  -  &         \\
  NL I3D + gcn   \cite{WangECCV2018}                   & R50           & 64=32$\times$2     &  46.1    &  76.8 & -    &  - &       \\
  CPNet\cite{LiuCVPR2019}                              & R34              & 2,304 =24$\times$16$\times$6    & -    & -& 57.65  & 83.95  &  N/A   \\
  TSM  \cite{LinICCV2019}                              & R50              & 48=8$\times$2$\times$3    & -    & -& 59.1  & 85.6  &      \\
  TSM  \cite{LinICCV2019}                              & R50              & 96=16$\times$2$\times$3   & -    & -& 63.4   &88.5  &      \\
 
%  STM   \cite{JiangICCV2019}                             & R50            & 240=8$\times$10$\times$3    & 49.2    & 79.3 & 62.3    & 88.8 &      \\
%  STM   \cite{JiangICCV2019}                             & R50            & 480=16$\times$10$\times$3   & 50.7    & 80.4 & 64.2    & 89.8  &       \\
  \hline
  Our SmallBig                        & R50            & 48=8$\times$2$\times$3    & 48.3   & 78.1  & 61.6  &  87.7   &       \\
  Our SmallBig                        & R50            & 96=16$\times$2$\times$3   & 50.0   & 79.8  & 63.8   & 88.9    &    N/A     \\
  Our SmallBig$_{En}$                 & R50            & 144=24$\times$2$\times$3  & \textbf{51.4}   & \textbf{80.7}  &  \textbf{64.5}    & \textbf{89.1}   &         \\
  \hline\hline
\end{tabular}
}
\caption{Comparisons with SOTA on Something-Something V1 and V2 validation set (RGB input).
For both V1 and V2,
our SmallBig-R50 achieves the best accuracy,
w.r.t.,
single-clip \& center-crop and multi-clip \& multi-crop.
Moreover,
our 8-frame SmallBig-R50 even outperforms 48-frame TSM-R50 for V2.
For multi-clip \& multi-crop,
the goal is to report the best accuracy.
Hence,
GFlops is not taken into account,
as suggested by TSM.}
\label{tab:kSSv2}
\vspace{-0.3cm}
\end{table}

\subsection{Evaluation on Something-Something V1 \& V2}

Due to lower resolution and shorter video length in Something-Something V1 and V2,
we adopt the slow-only baseline
\cite{FeichtenhoferICCV2019} 
for our SmallBig Net,
where
we add the SmallBig-Extra unit 
in Fig.\ref{fig:Blocks}(f)
respectively on top of res3, res4, and res5 stages of this baseline.  
Following \cite{LinICCV2019},
we group the results according to the number of sampled frames in the testing phase,
i.e.,
the single-clip \& center-crop case and the multi-clip \& multi-crop case.
For the multi-clip \& multi-crop case,
the goal is to report the best performance.
Hence,
GFlops is not taken into account,
as suggested in \cite{LinICCV2019}.
The results are shown in Table \ref{tab:kSSv2}.
For both V1 and V2,
our SmallBig-R50 achieves the best accuracy,
w.r.t.,
single-clip \& center-crop and multi-clip \& multi-crop.
%our 144-frame SmallBig-R50 achieves the best accuracy on V1.
%It achieves a competitive accuracy to 480-frame STM-R50 \cite{JiangICCV2019},
%with 3.3$\times$ less frames.
Moreover,
our 8-frame SmallBig-R50 even outperforms 48-frame TSM-R50 \cite{LinICCV2019} for V2 (Top1 acc: 59.7 vs. 59.1).
%Our 24-frame SmallBig-R50 even outperforms 240-frame STM-R50 \cite{JiangICCV2019} for V1 (Top1 acc: 50.4 vs. 49.2).
All these results further indicate that,
our SmallBig network can effectively boost video classification accuracy.

\begin{figure*}[t]
\centering
\includegraphics[width=1\textwidth]{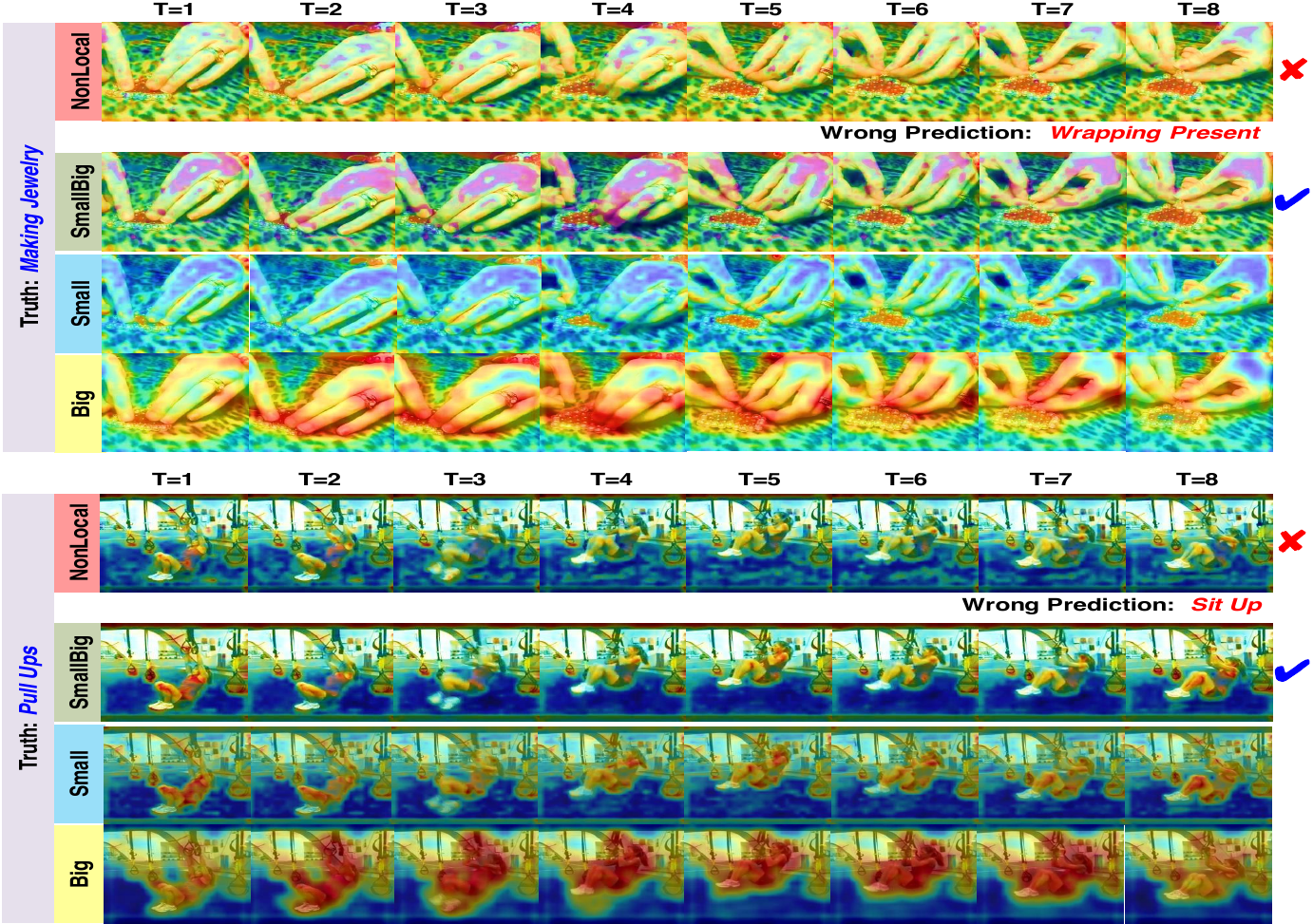}%[width=\textwidth]{Visualization2.pdf}%
\caption{Visualization.
Compared to the nonlocal network,
our SmallBigNet can discover key video details (e.g., \textit{Making Jewelry}) as well as reduce noisy backgrounds (e.g., \textit{Pull Ups}) for correct prediction.
More explanations can be found in Section \ref{sec:vis}.}
\label{fig:Visual}
\vspace{-0.2cm}
\end{figure*}

\subsection{Visualization}
\label{sec:vis}

We visualize and analyze the convolution features learned by SmallBigNet.
For comparison,
we use the nonlocal network \cite{Xiaolongcvpr2018} as a strong baseline.
Specifically,
we feed $8\times224\times224$ clips respectively into SmallBig-R23 and Nonlocal-R23,
and then extract $8\times28\times28$ convolution feature from res3.2 (after SmallBig and Nonlocal operations).
Finally,
we average the feature maps along the channel dimension,
and show them on the original image.
Fig.\ref{fig:Visual} clearly demonstrates that,
our SmallBig network can discover the key video details as well as reduce the noisy backgrounds,
compared to the nonlocal network. From this visualization, we also discover that the highly-activated points in the feature map distribute very sparsely in nonlocal, but ours can gather together.
This further validates our discussions in Section \ref{sec:discussion},
where
our SmallBig network is preferable to learn highly-activated contexts for video classification.

Furthermore,
we visualize small and big views from the first layer.
As expected,
small view tends to capture discriminative core semantics,
while big view tends to discover important contextual semantics.
For \textit{Making Jewelry},
small view captures hand contour and jewelry object,
while big view highlights the regions that contain key hand actions.
For \textit{Pull Ups},
small view captures key human parts and objects,
while big view highlights the most activated action regions.
By aggregating big contextual view to enhance small core view,
our SmallBig network is preferable to aggregate the core and contextual views for video classification and can effectively learn spatio-temporal representations.

\section{Conclusion}
In this work,
we propose a concise and novel SmallBig network with cooperation of small and big views.
In particular,
we enlarge the spatio-temporal receptive field in the big view branch,
in order to find the most activated context to enhance core representations in the small view branch.
Moreover,
we propose a parameter sharing scheme in our design,
which allows us to make the SmallBig network compact.
Finally,
all the experiments show that,
our SmallBig network is an accurate and efficient model for large-scale video classification.

\textbf{Acknowledge}
This work is partially supported by Science and Technology Service Network Initiative of Chinese Academy of Sciences (KFJ-STS-QYZX-092), Guangdong Special Support Program (2016TX03X276), and National Natural Science Foundation of China (61876176, U1713208), Shenzhen Basic Research Program (JCYJ20170818164704758, CXB201104220032A), the Joint Lab of CAS-HK, Shenzhen Institute of Artificial Intelligence and Robotics for Society.

{\small
\bibliographystyle{ieee_fullname}
\bibliography{egbib}
}

\end{document}